\documentclass[runningheads]{llncs}

\usepackage[year=2026]{accv}

\usepackage{accvabbrv}

\usepackage{graphicx}%
\usepackage{array}
\usepackage{multirow}%
\usepackage{amsmath,amssymb,amsfonts}%
\usepackage{mathrsfs}%
\usepackage[title]{appendix}%
\usepackage{xcolor}%
\usepackage{textcomp}%
\usepackage{manyfoot}%
\usepackage{booktabs}%
\usepackage{listings}%
\usepackage{subcaption} 
\usepackage{adjustbox}
\usepackage{tabularx}
\usepackage{latexsym}
\usepackage{epsfig}
\usepackage{epstopdf}
\usepackage{lscape}
\usepackage{mhsetup}
\usepackage{mathtools}
\usepackage{float}
\usepackage{xfrac}
\usepackage{pdfpages}
\usepackage[table]{xcolor}
\usepackage{colortbl}  

\usepackage[accsupp]{axessibility}  

\usepackage[pagebackref,breaklinks,colorlinks,citecolor=accvblue]{hyperref}

\usepackage{orcidlink}

\begin{document}

\title{UBLLIE: Unified Backlight and Low-Light Image Enhancement}


\author{Yasmin Yasin\inst{1}\orcidlink{0000-0000-0000-0000} \and
Muhammad Usman\inst{2}\orcidlink{0000-0003-2059-7206} \and
Ibrahim Radwan\inst{3}\orcidlink{0000-0002-8170-5058} \and
Saeed Anwar\inst{4}\orcidlink{0000-0002-0692-8411}}

\authorrunning{Yasin et al.}

\institute{King Fahd University of Petroleum and Minerals, Dhahran, Saudi Arabia \and
Ontario Tech University, Oshawa, Canada \and
University of Western Australia, Perth, Australia \and
University of Canberra, Canberra, Australia
}

\maketitle

\begin{abstract}
  Backlit and low-light images often suffer from severe exposure imbalance or global underexposure, presenting significant challenges for both visual perception and downstream computer vision tasks. In this paper, we propose a unified, unsupervised enhancement framework that addresses both types of degradation without relying on paired ground-truth data. Our approach builds on CLIP-guided prompt learning to semantically supervise enhancement using learned positive and negative textual prompts. To improve the quality of our improvements over prior work, we design a symmetric residual U-Net backbone augmented with an Atrous Spatial Pyramid Pooling module. This architecture captures multi-scale contextual information, enabling adaptive correction under spatially heterogeneous illumination. During training, the enhancement network is guided by CLIP-based semantic similarity losses and refined via an iterative prompt optimization mechanism. Extensive experiments on both paired and unpaired datasets, including BAID, Backlit300, LOL, and VE-LOL-L, demonstrate that our framework consistently outperforms state-of-the-art supervised and unsupervised methods in terms of fidelity, perceptual quality, and generalization. Furthermore, our work emphasizes the need for stronger benchmarking protocols for backlit enhancement, a relatively underexplored area. The proposed framework provides a robust, scalable solution for real-world illumination enhancement across diverse lighting conditions.
\end{abstract}

\section{Introduction}\label{sec1}

Backlit images, in which the principal light source is behind the objects, result in significant illumination imbalances. This phenomenon results in areas that are either excessively bright or overly dark~\cite{Liang_2023_ICCV}. Such uneven lighting not only degrades image quality but also degrades the performance of essential computer vision algorithms, including object recognition, edge detection, and tracking~\cite{guo2016lime}. Enhancing backlit images is critical for both aesthetic and functional purposes, ensuring clearer images that enable vision-based applications~\cite{tian2023survey} to perform accurately across fields such as autonomous driving~\cite{li2021low}, medical imaging, and security surveillance~\cite{9063535}. Poorly lit images degrade the performance of these algorithms, leading to inefficiencies and errors~\cite{9190929}.

\begin{figure*}[t]
\centering
\begin{tabular}{cccccccc}
\includegraphics[width=0.115\linewidth]{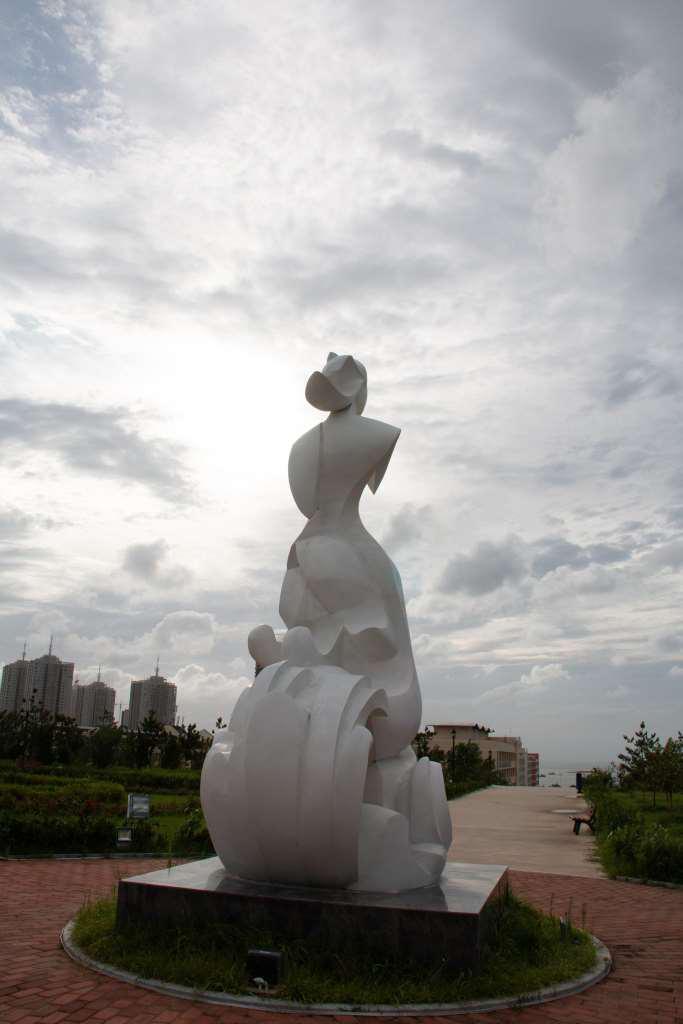}&
\includegraphics[width=0.115\linewidth]{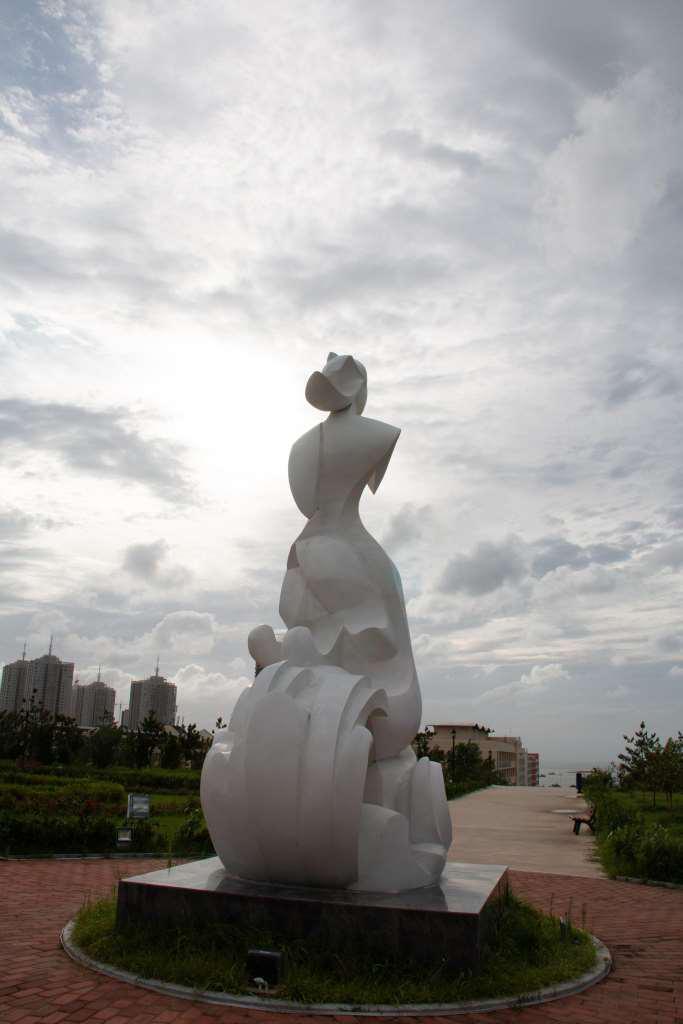}& 
\includegraphics[width=0.115\linewidth]{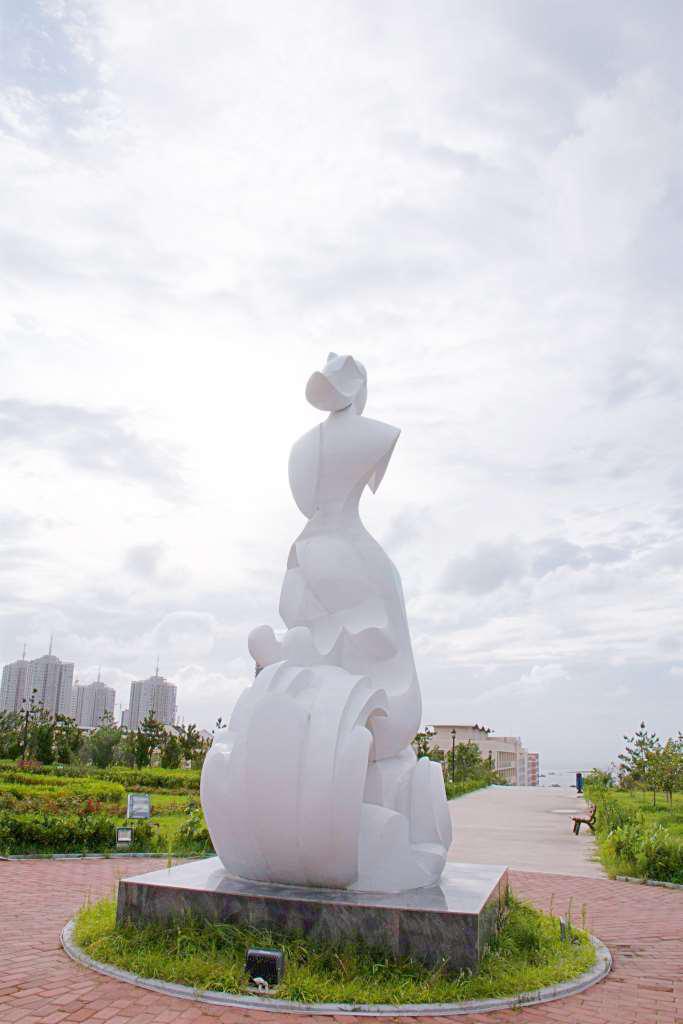} &
\includegraphics[width=0.115\linewidth]{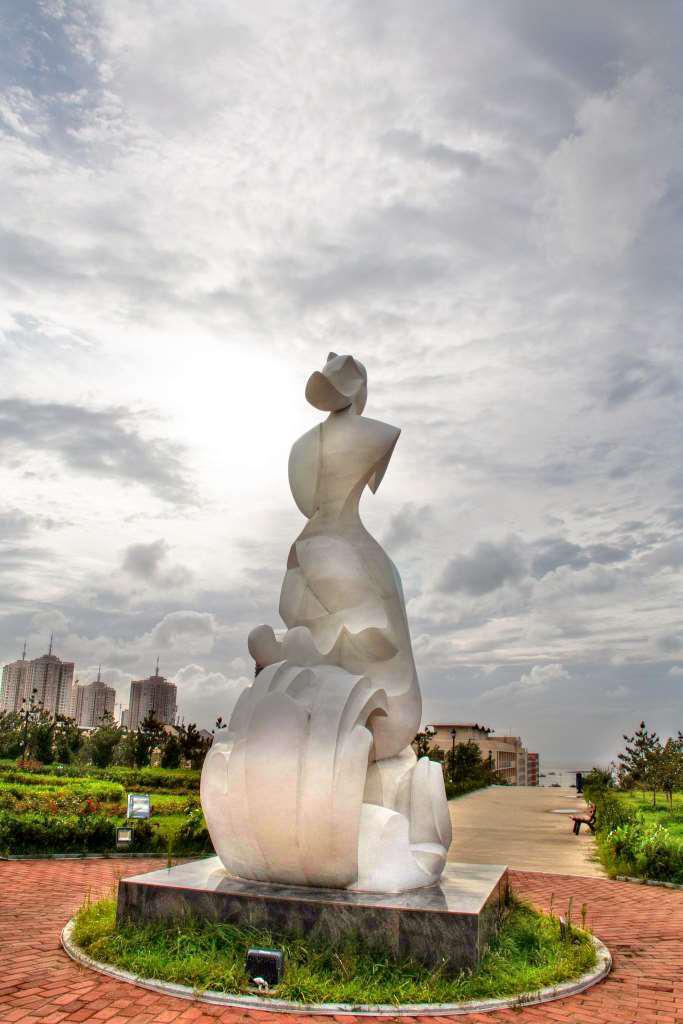}&
\includegraphics[width=0.115\linewidth]{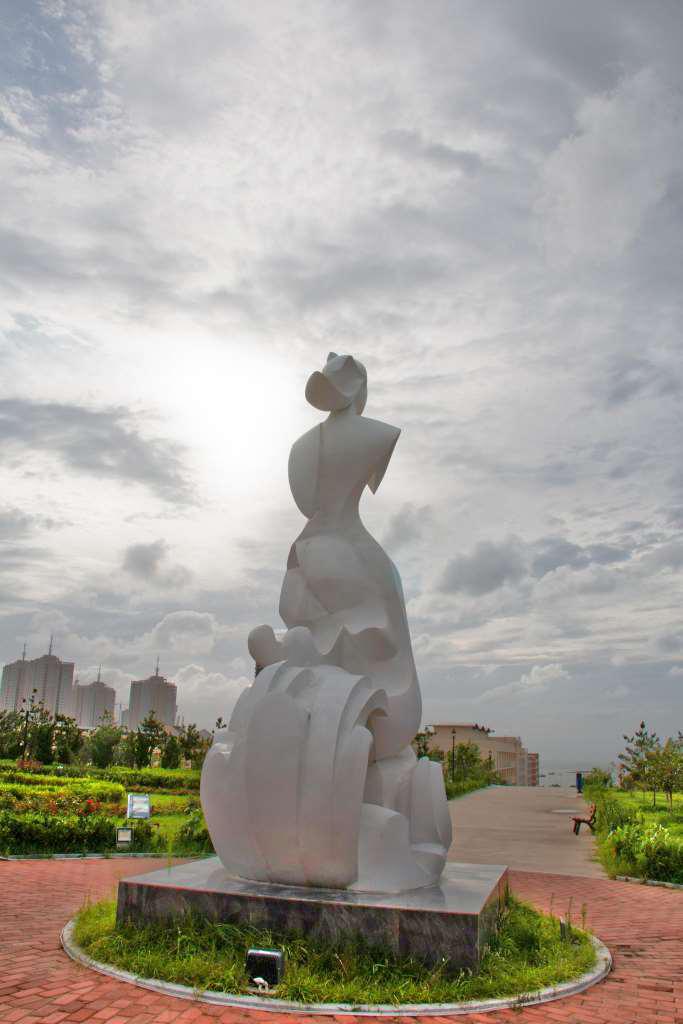}&
\includegraphics[width=0.115\linewidth]{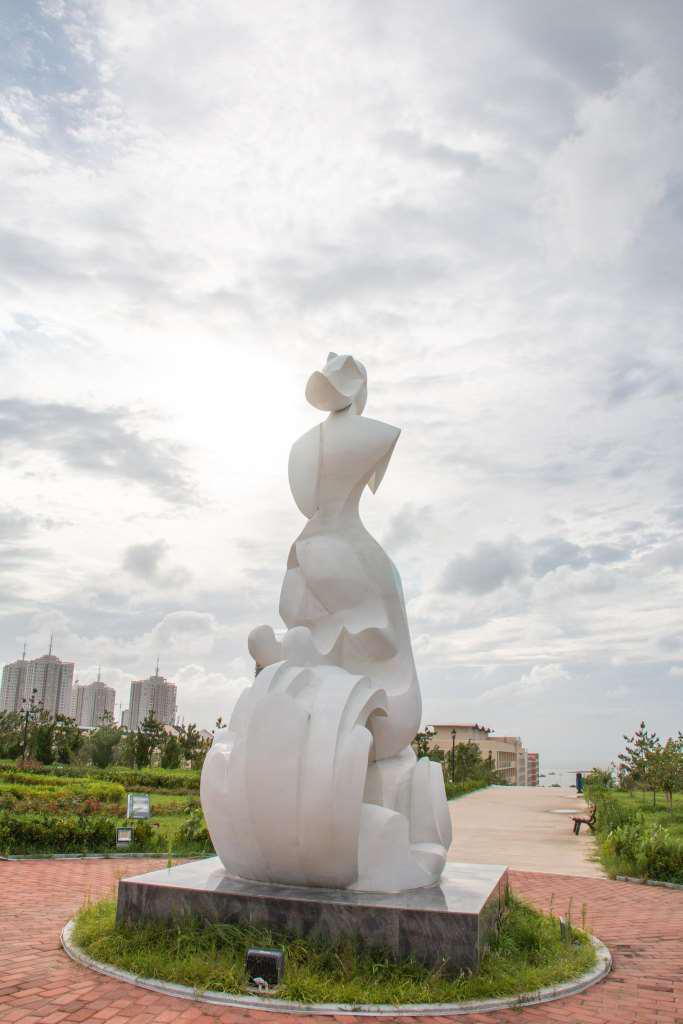}& 
\includegraphics[width=0.115\linewidth]{images/Clip.jpg}& 
\includegraphics[width=0.115\linewidth]{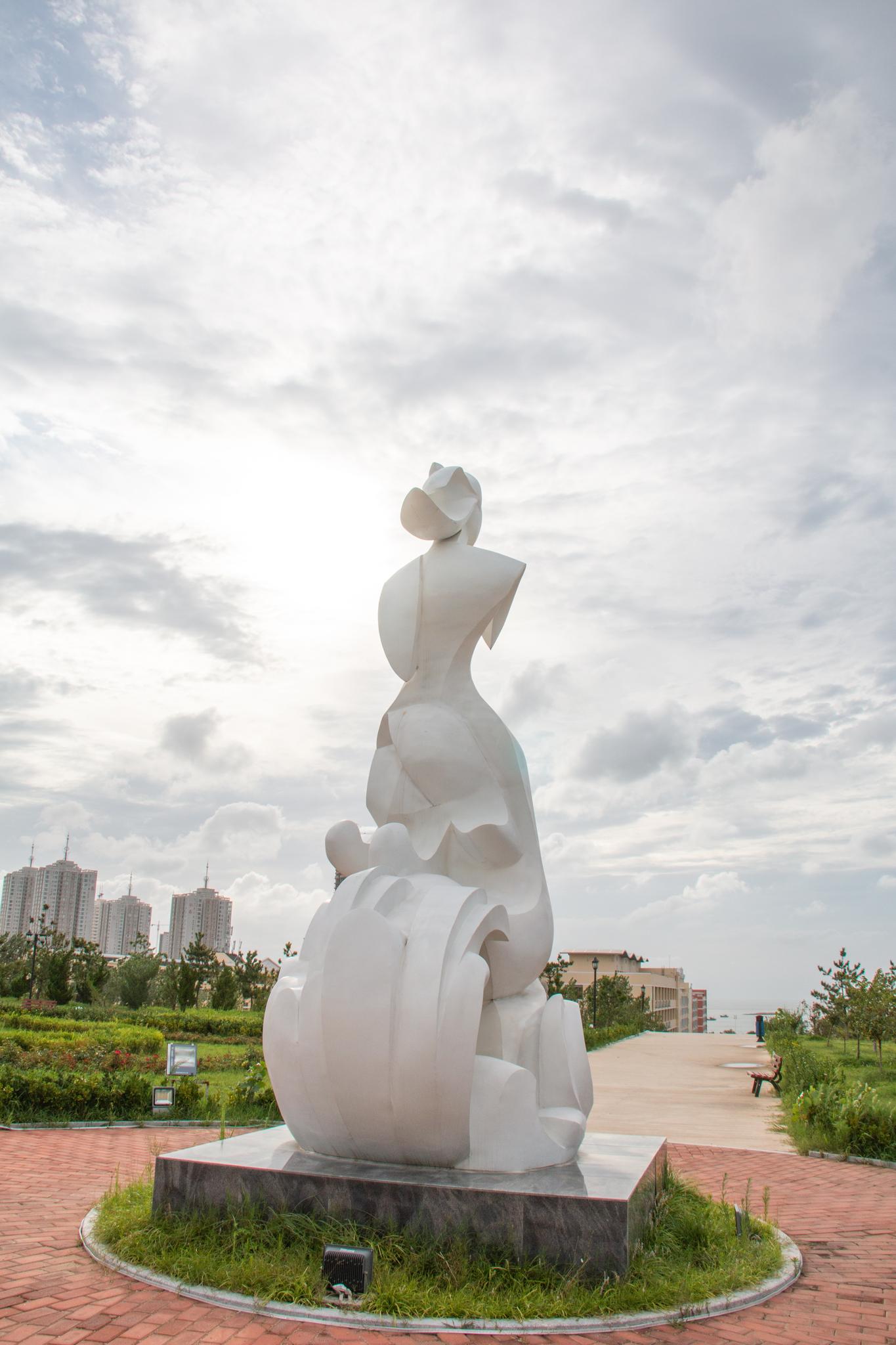}\\

Input & ~\cite{afifi2021learning} &~\cite{guo2020zero} &~\cite{jiang2021enlightengan} &~\cite{zhang2019zero} &~\cite{Liang_2023_ICCV} & Ours & GT\\
\end{tabular}
\caption{Visual comparison on backlit images sampled from the BAID test dataset. }
\label{fig:backlit_baid}
\end{figure*}

Traditional methods, including Histogram Equalization (HE) and Retinex-based techniques~\cite{fan2023lacn,10299637}, such as SSR~\cite{jobson1997properties} and MSRCR~\cite{jobson1997multiscale}, have been employed to enhance backlit images by adjusting contrast and separating illumination from reflection components. However, these approaches often face challenges, including over-enhancement, complex lighting conditions, and difficulty balancing background and foreground exposure~\cite{xue2024low}. Consequently, images may appear artificial, and dynamic range compression in backlit photographs remains a significant challenge. The manual correction of such photos is labor-intensive, as it necessitates the enhancement of underexposed regions while maintaining the quality of well-illuminated areas~\cite{Liang_2023_ICCV}. Despite the availability of automated enhancement techniques~\cite {xue2024low}, these methods face significant challenges in addressing backlit conditions~\cite{li2021low}. The principal challenge is the inherently ill-posed nature of disentangling illumination components from a single image, which often yields suboptimal results~\cite{lv2022backlitnet}. Additionally, numerous learning-based approaches rely on supervised training, thereby introducing biases through subjectively curated expert ground-truth data~\cite{xue2024low}. Such biases undermine the real-world effectiveness of enhancement models.

In addition to backlit conditions, low-light images—characterized by globally insufficient illumination—pose similar challenges for both human perception and automated analysis. Low-light scenes commonly occur in indoor environments or at night, where insufficient exposure leads to suppressed detail, reduced contrast, and increased noise. These degradations not only obscure critical visual information but also hinder downstream tasks such as detection, segmentation, and tracking. Compared with backlit scenes affected by localized exposure imbalances, low-light images generally exhibit uniform darkness, thereby complicating traditional enhancement techniques that rely on spatially varying illumination. Moreover, obtaining paired low- and normal-light data for training is both costly and time-consuming, making low-light enhancement particularly suitable for unsupervised or zero-shot learning paradigms. Effective enhancement must not only brighten the image but also restore texture, mitigate noise, and preserve natural color fidelity—goals that require both a global and local understanding of scene illumination~\cite{he2026unfoldir}.

In response to these challenges, deep learning has emerged as a powerful solution for improving the quality of backlit and low-light images. Leveraging large datasets and sophisticated neural network architectures, recent techniques such as Generative Adversarial Networks (GANs)~\cite{wang2022magan,tang2023improved,xu2023low}, Diffusion Models~\cite{jiang2023low,wu2023reco,zhu2023denoising}, Contrastive Language–Image Pretraining (CLIP)~\cite{Zheng_2022_WACV,Yang_2023_ICCV,wang2024dap}, Transformers~\cite{cai2023retinexformer,cui2022tpet,yang2023lightingnet}, and U-Net~\cite{hai2023r2rnet,shi2024ll,park2024dedu} facilitate the attainment of more accurate and visually natural outcomes. These models can adapt to diverse lighting conditions and the complexities of various scenes~\cite{creswell2018generative}, thereby producing enhancements well-suited for real-time applications in domains such as autonomous driving and surveillance. Among the different learning paradigms, supervised~\cite{jiang2023low,wu2023reco,zhou2023pyramid}, unsupervised~\cite{morawski2024unsupervised,Liang_2023_ICCV}, and zero-shot~\cite{Zheng_2022_WACV,tian2023zero} learning are extensively employed, with an increasing inclination towards unsupervised and zero-shot methods. These approaches enable models to generalize more effectively by leveraging unpaired data, thereby capturing the variability of real-world conditions without requiring manually labeled paired datasets. Despite advances in deep learning for backlit and low-light image enhancement, challenges remain, including the need for large datasets and computational resources, as well as the critical importance of preserving naturalistic results without artifacts. Future efforts will focus on developing efficient algorithms and diverse datasets to enhance robustness and generalization, while also integrating traditional methods with deep learning approaches~\cite{li2021low}.

\textbf{Contributions:}
We present a unified, unsupervised framework for enhancing images in both backlight and low-light conditions, without relying on paired data. Our methodology incorporates CLIP-guided prompt learning to deliver high-level semantic supervision via learned textual prompts. In contrast to previous works, we augment the image-to-image architecture by integrating a symmetric residual U-Net equipped with an Atrous Spatial Pyramid Pooling (ASPP) module. This enhancement enables the network to effectively capture multi-scale illumination features and accommodate spatially heterogeneous exposure conditions. We demonstrate the generalizability of our framework across various domains, showcasing robust performance on both paired and unpaired datasets. Moreover, we highlight the need for standardized benchmarking for backlit enhancement and provide robust baselines on less-explored datasets, such as BAID and Backlit300.

\section{Related Works}
\label{sec:related}
The domain of image enhancement has seen considerable advances, predominantly in low-light image enhancement (LLIE~\cite{Hua_2026_CVPR,Xu_2026_CVPR,Cho_2026_CVPR}), whereas backlit image enhancement (BLIE)~\cite{wu2026control,gaintseva2024rave} has received less attention. Techniques employed for LLIE, including deep learning models such as GANs, Contrastive Language-Image Pretraining (CLIP), diffusion models, and U-Net, can also be adapted for BLIE. Nevertheless, the principal distinction lies in the distribution of light: LLIE pertains to uniformly underexposed areas, whereas BLIE encompasses pronounced imbalances between luminous and shadowy regions, rendering it more complex. In this research, our focus is on BLIE as a method adept at addressing its intricate lighting imbalances, which is expected to perform effectively in LLIE, where light distribution is generally more homogeneous. Once established, the BLIE technique will be evaluated and applied within LLIE tasks.

\vspace{1mm}\noindent\textbf{GANs Based BLIE and LLIE.} GAN-based architectures, such as LAE-Net~\cite{liu2023lae}, DRGN~\cite{jiang2022degrade}, and LumiNet~\cite{bose2023luminet}, have illustrated considerable advancements in the field. LAE-Net employs repeated generator blocks to produce multi-scale feature maps, while both local and global discriminators assess image quality. Its kernel-selection method, known as EIKS, along with illumination-attentive transfer subnets, enhances feature extraction and increases the visibility of noise. DRGN comprises a degradation generator, a refinement generator, and a multi-resolution fusion network that decomposes learning tasks into subspaces and refines the representation of multi-scale features. LumiNet adopts a multispatial attention-based U-Net as its generator, 
and uses a patch-based GAN discriminator to facilitate stable training.

\vspace{1mm}\noindent\textbf{CLIP Based BLIE and LLIE.} CLIP, a model introduced by OpenAI~\cite{radford2021learning}, has been leveraged for BLIE by aligning text prompts with visual content. In CLIP-LIT~\cite{Liang_2023_ICCV}, a two-stage process refines text-image alignment through prompt learning and rank learning, which enhances backlit images without modifying the CLIP model. Similarly, the CFWD~\cite{xue2024low} framework combines CLIP with wavelet and Fourier transforms to improve structural similarity and restore backlit images. Morawski~et~al.~\cite{morawski2024unsupervised} proposed a two-stage training process that uses CLIP for semantic guidance to enhance low-light images without using paired or unpaired data, thereby rendering it adaptable to diverse datasets. These approaches showcase CLIP's effectiveness in image enhancement tasks.

\vspace{1mm}\noindent\textbf{Diffusion Models Based BLIE and LLIE.} Diffusion models have gained prominence for image enhancement and generation, preserving details while addressing backlighting. CFWD~\cite{xue2024low} employs wavelet transforms to systematically decompose images, apply diffusion operations, and calculate text-image similarity, thereby enhancing structural similarity through a hybrid approach. WCDM~\cite{jiang2023low} enhances inference speed and minimizes resource consumption through the utilization of wavelet transformations, ensuring stability by combining forward diffusion and denoising during training, along with a high-frequency restoration module aimed at the enhancement of fine details. PyDiff~\cite{zhou2023pyramid} presents a pyramid diffusion technique that accelerates sampling without compromising performance and incorporates a global corrector to reduce degradation. DiffLight~\cite{feng2024difflight} integrates a Denoising Enhancement branch dedicated to noise reduction and brightness enhancement, complemented by a Detail Preservation branch which employs a Light Full-Former method to maintain image details. GDP~\cite{fei2023generative} uses a pre-trained denoising diffusion generative model to address linear and nonlinear challenges, optimizing its parameters for effective blind image restoration.

\vspace{1mm}\noindent\textbf{U-Net Based BLIE and LLIE.} U-Net architectures are central to BLIE due to their efficient feature extraction and reconstruction capabilities. In models such as SGZ~\cite{zheng2022semantic}, BacklitNet~\cite{lv2022backlitnet}, and LumiNet~\cite{bose2023luminet}, U-Net-inspired designs tackle the complexities of backlit scenes by capturing both local and global contexts. SGZ presents the Enhancement Factor Extraction Network specifically for feature extraction. In contrast, BacklitNet employs a dual-resolution framework with a nested U-shaped structure to enhance backlit images by extracting multi-scale features. LumiNet implements a multispatial attention-based U-Net to generate enhanced images with increased detail and a natural appearance, thereby highlighting the adaptability of U-Net architectures for image enhancement.

In addition to the earlier models, several others have been proposed for BLIE and LLIE. LLHFNet~\cite{al2022low} extracts high-level features from the Value channel of HSV images using networks such as VGG16 or ResNet50, thereby generating filter parameters for enhancement. LACN~\cite{fan2023lacn} employs ConvNeXt, incorporating modules specifically designed for noise suppression and illumination evaluation. Bread-ME~\cite{guo2023low} features sub-networks aimed at illumination adjustment, noise suppression, and color adaptation, while LLFlow~\cite{wang2022low} utilizes an encoder-decoder architecture with Residual-in-Residual Dense Blocks for color mapping. LLFormer~\cite{wang2023ultra} combines hierarchical structures with transformer blocks for effective feature extraction. U2E-Net~\cite{khan2024lit} divides images into reflection and illumination components, and UDCN~\cite{jiang2022unsupervised} includes sub-networks for comprehensive image decomposition and correction. URetinex-Net~\cite{wu2022uretinex} decomposes images into reflectance and illumination, thereby highlighting diverse deep learning approaches for BLIE and LLIE.

\section{Methodology}
\begin{figure}[t]
\centering
\begin{tabular}{c}
        \includegraphics[width=0.75\linewidth]{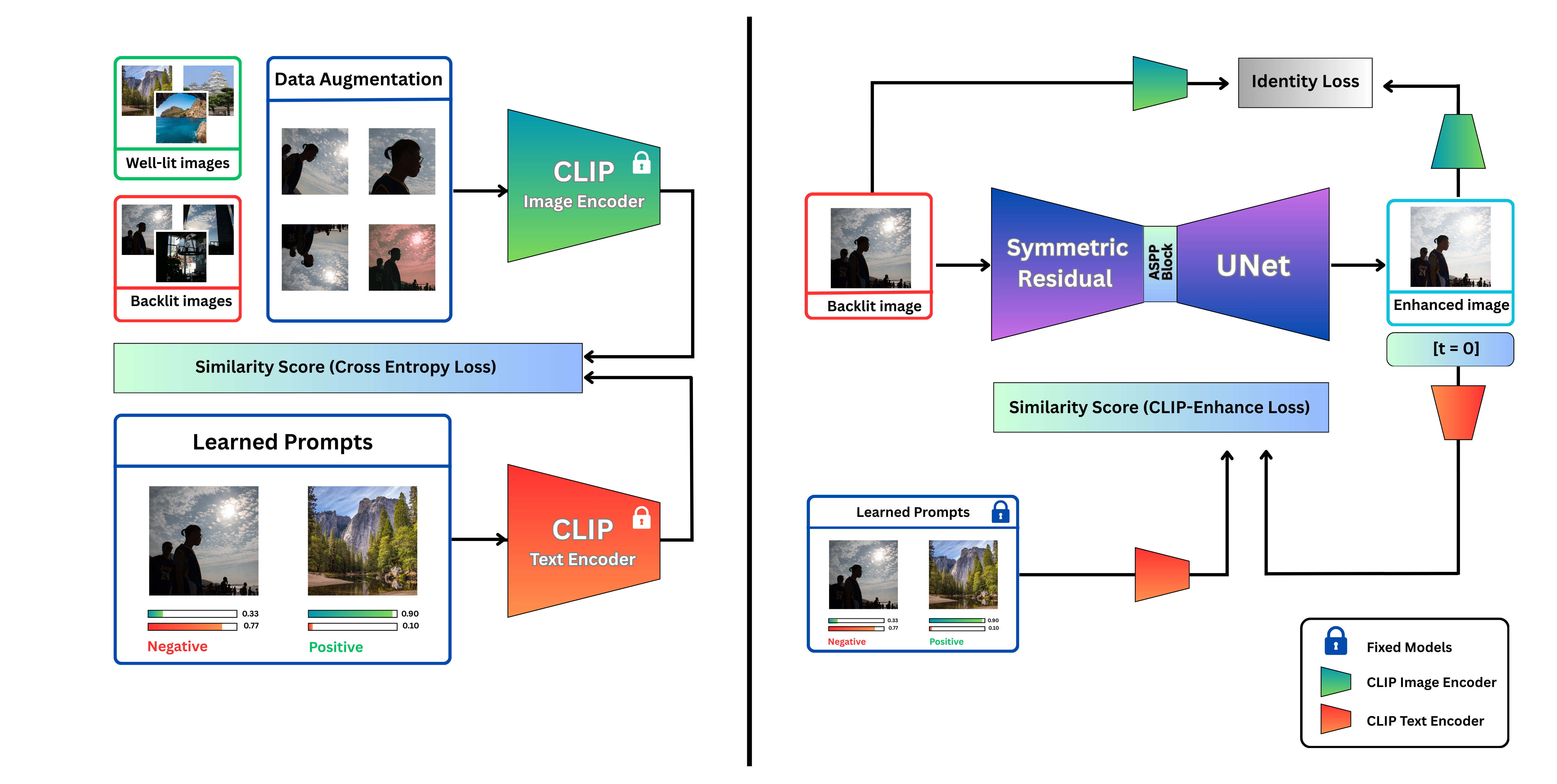}\\
        a) Prompt Initialization and Initial Enhancement Network Training\\
        
        \includegraphics[width=0.7\linewidth]{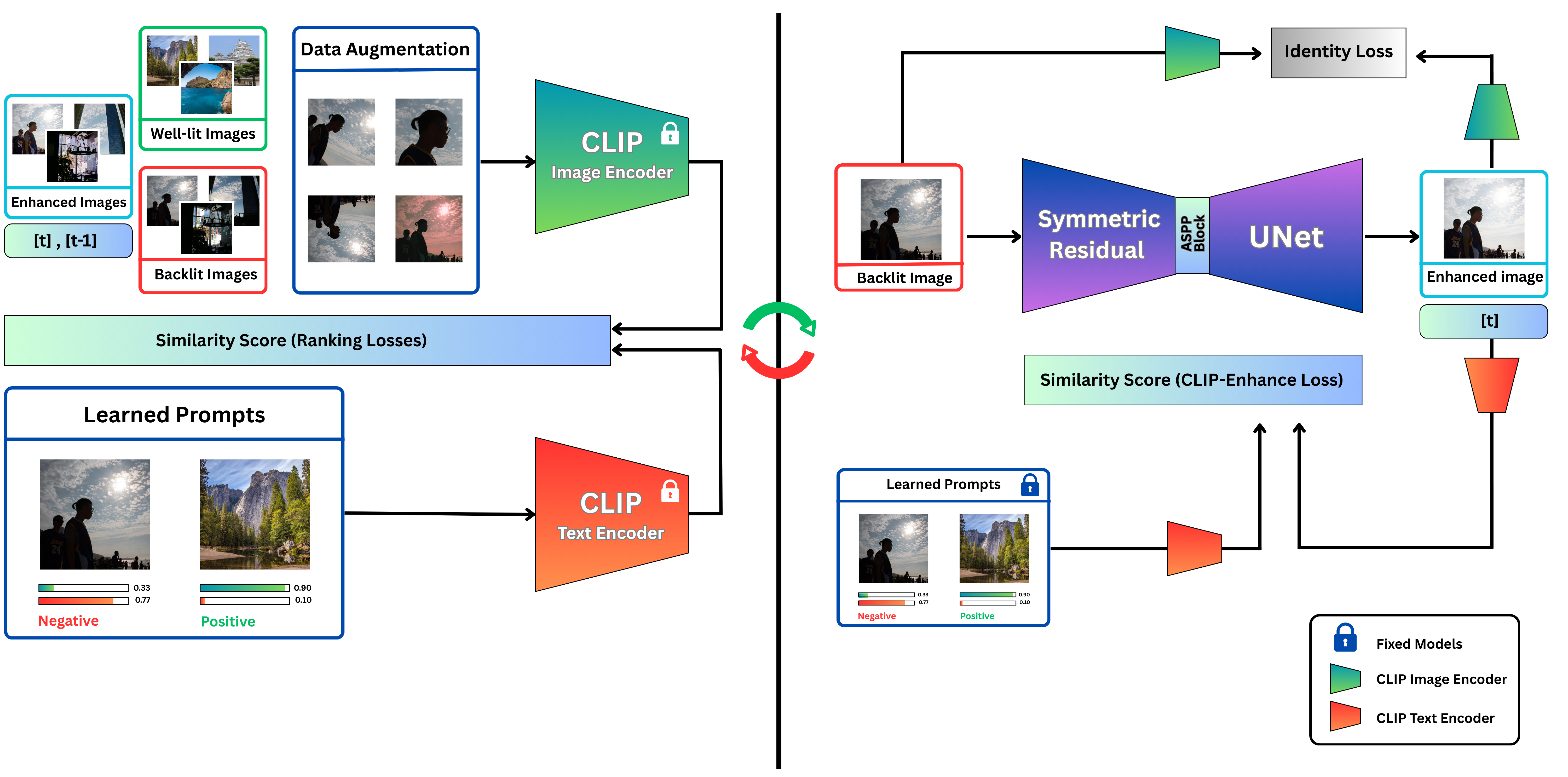}\\
        b) Prompt Refinement and Enhancement Model Fine-tuning
\end{tabular}



\caption{Overview of our proposed unified enhancement framework. (a) During the first stage, prompts for both well-lit and poorly lit conditions are initialized using CLIP similarity scores, and a preliminary enhancement network is trained to align its outputs with the positive prompt. (b) In the second stage, the prompts are refined using a ranking loss that compares the input, enhanced, and reference image embeddings. The enhancement network is then fine-tuned using the updated prompts. The enhancement backbone comprises a symmetric residual U-Net integrated with an ASPP module to effectively capture multi-scale illumination context.}
\label{fig:method}
\end{figure}

Our proposed framework addresses the unified task of enhancing both backlight and lowlight images by integrating semantic guidance from CLIP with a carefully designed enhancement network. As illustrated in Figure~\ref{fig:method}, the framework operates in four main stages: prompt initialization, initial enhancement training, prompt refinement, and final enhancement model fine-tuning. During training, we learn a pair of prompts that represent well-lit and poorly lit visual concepts, which guide the enhancement network via CLIP-based supervision, following the general strategy introduced in CLIP-LIT~\cite{Liang_2023_ICCV}. Unlike previous works, we significantly enhance the network backbone by introducing a symmetric residual U-Net architecture equipped with an ASPP module~\cite{chen2017deeplab}. This architectural improvement enables the model to capture multi-scale illumination cues and spatial context more effectively, enhancing robustness across varied lighting conditions. Together, the semantic supervision and structural improvements enable our framework to generalize well across both backlit and lowlight images, without requiring paired training data.

\subsection{Prompt Initialization}
\label{subsec:prompt-init}

The initial phase of our framework focuses on learning a pair of textual prompts that semantically represent well-lit and poorly-lit image conditions. These prompts, designated as the positive prompt $P^+$ and the negative prompt $P^-$, direct the enhancement network through CLIP-based supervision~\cite{radford2021learning}. We establish two reference image sets: one containing well-exposed images and another containing degraded photos, such as low-light and backlit images. Each image is embedded using CLIP's frozen image encoder, while the prompt tokens are passed through the text encoder to obtain their respective textual embeddings.

To optimize the prompts, we minimize the cross-entropy loss with respect to the cosine similarities between the image embeddings and both textual prompts. For a given reference image $I$, the loss is defined as:
\begin{equation}
\mathcal{L}_{\text{init}} = -\log \frac{e^{\text{sim}(E_I(I), E_T(P^+))}}{e^{\text{sim}(E_I(I), E_T(P^+))} + e^{\text{sim}(E_I(I), E_T(P^-))}},
\label{eq:init-loss}
\end{equation}
where $E_I(\cdot)$ and $E_T(\cdot)$ are the image and text encoders of CLIP, respectively, and $\text{sim}(\cdot)$ denotes cosine similarity. Well-lit images are paired with $P^+$, whereas low-quality images are associated with $P^-$ during the optimization process. This contrastive training facilitates the positive prompt to represent well-lit semantics, while the negative prompt captures illumination degradation.

The initialized prompts $\{P^+, P^-\}$ serve as soft semantic anchors within the CLIP space. Subsequently, they are used to compute the enhancement loss, which guides the enhancement model to produce outputs semantically aligned with the well-lit domain. This initialization strategy adheres to the conventional method employed in CLIP-LIT~\cite{Liang_2023_ICCV}, but is modified in our framework to support the dual-domain setting encompassing both low-light and backlit degradation.

\subsection{Initial Enhancement Network Training}
\label{subsec:initial-training}

With the learned prompts in place, we train the image enhancement network with CLIP-guided semantic supervision. At this stage, the goal is to optimize the network to generate enhanced outputs that are semantically aligned with the positive prompt $P^+$ while being distinct from the negative prompt $P^-$. It is essential to note that this training process operates independently of any paired ground-truth images, thereby categorizing it as entirely unsupervised.

Given an input image $I_{\text{in}}$ sourced from a low-light or backlit dataset, the enhancement network $G(\cdot)$ generates an output $I_{\text{out}} = G(I_{\text{in}})$. The input and output are embedded using the CLIP image encoder, and their similarity to the learned textual prompts is computed. The network is supervised using the same contrastive formulation as during the prompt initialization phase, now applied to the enhanced outputs. The CLIP loss is expressed as follows:
\begin{equation}
\mathcal{L}_{\text{clip}} = -\log \frac{e^{\text{sim}(E_I(I_{\text{out}}), E_T(P^+))}}{e^{\text{sim}(E_I(I_{\text{out}}), E_T(P^+))} + e^{\text{sim}(E_I(I_{\text{out}}), E_T(P^-))}},
\label{eq:clip-loss}
\end{equation}
where $\text{sim}(\cdot)$ denotes cosine similarity, and $E_I(\cdot)$ and $E_T(\cdot)$ are CLIP's image and text encoders. This loss encourages an enhanced image toward the well-lit semantic space defined by $P^+$ and away from the poorly lit representation $P^-$. Given that the supervision signal is derived entirely from pretrained CLIP embeddings and learned prompts, this methodology enables effective training without explicit paired data. During this phase, the enhancement network gradually learns to correct global illumination imbalances, enhance darker regions, and preserve well-lit areas. The quality of this preliminary training lays the foundation for subsequent stages of prompt refinement and model fine-tuning.

\subsection{Prompt Refinement}
\label{subsec:prompt-refinement}

While the initial prompts $\{P^+, P^-\}$ serve a helpful starting point for semantic supervision, they may not comprehensively capture the nuances of lighting quality across diverse scenes. To address this, we refine the prompts utilizing a ranking-based loss that enhances their discriminative power and alignment with the enhancement task.

The refinement process considers the relationships among three image types: the original input $I_{\text{in}}$, the enhanced output $I_{\text{out}}$, and a well-lit reference image $I_{\text{ref}}$. These images are processed through the frozen CLIP image encoder to obtain their corresponding embeddings. The objective is to ensure that the similarity between $I_{\text{out}}$ and the positive prompt $P^+$ exceeds that of $I_{\text{in}}$, while also promoting $I_{\text{ref}}$ to be even closer to $P^+$ than $I_{\text{out}}$. To achieve this, we employ a soft ranking loss defined as:
\begin{align}
\mathcal{L}_{\text{rank}} &= \log\left(1 + e^{\text{sim}(E_I(I_{\text{in}}), E_T(P^+)) - \text{sim}(E_I(I_{\text{out}}), E_T(P^+))} \right) \\&+ 
\log\left(1 + e^{\text{sim}(E_I(I_{\text{out}}), E_T(P^+)) - \text{sim}(E_I(I_{\text{ref}}), E_T(P^+))} \right),
\end{align}
\label{eq:rank-loss}
where $E_I$ and $E_T$ are the CLIP encoders and $\text{sim}(\cdot)$ denotes cosine similarity.

By minimizing the ranking loss, the prompts are iteratively updated to better distinguish among poorly lit, enhanced, and well-lit images in the CLIP embedding space. This refined semantic representation is then used to supervise a second round of enhancement training, thereby improving perceptual quality and generalization. Prompt refinement enables the supervision signal to evolve as the enhancement network progresses, making the semantic guidance increasingly specific to the task at hand. This dynamic interaction between prompt optimization and image enhancement constitutes the foundation for the final stage of our framework.

\subsection{Enhancement Network Fine-Tuning}
\label{subsec:enhancement-finetune}

Upon refining the prompts, we retrain the enhancement network using the updated prompts $\{P^+, P^-\}$ as supervisory targets. This stage, illustrated in the second half of Fig.~\ref{fig:method}, builds upon the initial training by leveraging the enhanced semantic alignment of the prompts to provide more accurate and perceptually meaningful guidance.

The training methodology remains unsupervised and uses the same loss formulation as in the preliminary enhancement stage. For each input image $I_{\text{in}}$, the network generates an output $I_{\text{out}} = G(I_{\text{in}})$, and the CLIP-based contrastive loss is calculated using the refined prompts:

\begin{equation}
\mathcal{L}_{\text{clip}}^{\text{refined}} = -\log \frac{e^{\text{sim}(E_I(I_{\text{out}}), E_T(P^+))}}{e^{\text{sim}(E_I(I_{\text{out}}), E_T(P^+))} + e^{\text{sim}(E_I(I_{\text{out}}), E_T(P^-))}}.
\label{eq:clip-finetune}
\end{equation}

To maintain structural consistency with the input, we again include an identity loss term defined as:
\begin{equation}
\mathcal{L}_{\text{id}} = \| I_{\text{out}} - I_{\text{in}} \|_1,
\label{eq:identity}
\end{equation}
and the overall training objective becomes:
\begin{equation}
\mathcal{L}_{\text{total}} = \mathcal{L}_{\text{clip}}^{\text{refined}} + \lambda \mathcal{L}_{\text{id}},
\end{equation}
where $\lambda$ is a balancing coefficient; in our experiments, we set $\lambda = 0.9$ to assign equal importance to perceptual quality and content preservation, consistent with the original setting in CLIP-LIT~\cite{Liang_2023_ICCV}.

This fine-tuning phase ensures that the network aligns not only with the refined semantic supervision but also becomes more robust against a broader range of lighting conditions. It allows the model to learn more nuanced mappings between degraded and high-quality visual domains, ultimately yielding outputs with improved perceptual and structural fidelity.

\subsection{Enhancement Network Architecture}
\label{subsec:enhancement-arch}

To effectively handle both backlit and low-light image enhancement, we design a specialized image-to-image network that combines the strengths of a U-Net backbone, residual learning, and Atrous Spatial Pyramid Pooling (ASPP). This architecture is tailored to address spatially heterogeneous lighting, a common feature in real-world scenes, where regions of significant underexposure and overexposure often coexist. We have selected the U-Net~\cite{Ronneberger2015UNet} architecture as the foundation of our enhancement model due to its established ability to preserve spatial information while performing hierarchical feature extraction. Initially proposed for biomedical segmentation, the U-Net proves particularly effective in contexts that involve pixel-level transformations alongside structured spatial priors. Within our framework, backlit and low-light images often exhibit sharp transitions between bright and dark regions, analogous to object boundaries in segmentation tasks. The encoder-decoder architecture of U-Net, further augmented by skip connections, enables the model to optimize global illumination correction while preserving local detail, which is critical for enhancing shadowed regions without excessively amplifying well-lit areas.

To further improve training stability and enhancement precision, we embed residual blocks~\cite{He2016ResNet} throughout both the encoder and the decoder. Each residual block comprises two $3 \times 3$ convolutional layers with ReLU activation, with a skip connection that adds the input to the output. Residual learning encourages the network to focus on learning the discrepancy between the input and the desired output, which is ideal for enhancement tasks in which the structure should be preserved and only illumination or contrast needs adjustment. This is particularly useful when handling images with mixed exposure, as it enables the model to apply changes selectively rather than uniformly.

A principal contribution of our architectural design is the inclusion of an Atrous Spatial Pyramid Pooling (ASPP) module~\cite{chen2017deeplab} inserted after the second encoder stage. ASPP comprises parallel atrous convolutions with dilation rates of 1, 6, 12, and 18, along with a global average pooling path. These features are concatenated and projected to form a rich multi-scale representation:
\begin{equation}
F_{\text{ASPP}} = [\alpha_{r=1}(f), \alpha_{r=6}(f), \alpha_{r=12}(f), \alpha_{r=18}(f), \rho(f)],
\end{equation}
where $\alpha_{r=k}$ denotes atrous convolution with a rate of $k$, and $\rho$ indicates global average pooling succeeded by a $1 \times 1$ convolution. This module allows the network to perceive both fine local structures and broader contextual cues, which is critical for adjusting illumination without introducing artifacts such as halos or over-saturation. In backlit scenes, where contrast fluctuates significantly across the image, and in low-light conditions, where textures are suppressed by darkness, ASPP provides the receptive-field diversity needed to balance exposure correction with structure preservation.

The entire architecture comprises an input convolutional layer, three encoder blocks, the ASPP module, three decoder blocks featuring upsampling and skip connections, and a final reconstruction layer. Each phase benefits from a residual design, while the ASPP module enhances context at the network's bottleneck. This integrated design enables the network to distinguish between globally and locally degraded regions, apply appropriate corrections, and generate perceptually enhanced outputs with high structural and tonal fidelity. Compared with the original CLIP-LIT~\cite{Liang_2023_ICCV} U-Net backbone, our model demonstrates superior capability to recover fine details and maintain balanced lighting across complex visual scenes.

\section{Experimental Setup}\label{sec2}
\vspace{2mm}\noindent\textbf{Datasets}. 
For training, we use the backlit subset of the BAID dataset~\cite{lv2022backlitnet}, comprising 380 paired images. Although ground-truth (GT) annotations are available, they are excluded during training to preserve a fully unsupervised learning setting. We also incorporate 384 well-lit images from the DIV2K dataset~\cite{agustsson2017ntire}, which serve as high-quality reference exemplars for prompt initialization and refinement. To address low-light scenarios, we use 400 paired images from the LOL dataset~\cite{wei2018deep}. 
To improve training diversity, we include 2,100 paired images from VE-LOL-L, which covers both synthetic and real low-light scenes~\cite{liu2021benchmarking}. For testing, we evaluate under backlit, low-light, and mixed lighting conditions using the BAID test split~\cite{lv2022backlitnet}, Backlit300~\cite{Liang_2023_ICCV}, LOL~\cite{wei2018deep}, and VE-LOL-L test images~\cite{liu2021benchmarking}. Backlit300 images are resized so that the longest side is at most 2048 pixels.

\begin{table}[t]

\caption{Quantitative comparison on the BAID~\cite{lv2022backlitnet} and Backlit300~\cite{Liang_2023_ICCV} test dataset. The best and second performances are marked in \textcolor{red}{\textbf{red}} and \textcolor{blue}{\underline{blue}}.}
\centering
\resizebox{0.85\textwidth}{!}{ 
\begin{tabular}{c|c|c c c c|| c}
            \hline \hline
             &  & \multicolumn{4}{c||}{BAID~\cite{lv2022backlitnet}}&Backlit300~\cite{Liang_2023_ICCV}\\
            
            Learning & Methods & PSNR $\uparrow$ & SSIM $\uparrow$ & LPIPS $\downarrow$ & MUSIQ $\uparrow$ & MUSIQ $\uparrow$ \\
            \hline\hline
            & Input & 16.641 & 0.768 & 0.197 & 52.115 & 51.900 \\\hline 
            
            \multirow{7}{*}{Supervised} 
            & Afifi~et~al.\cite{afifi2021learning} & 15.904 & 0.745 & 0.227 & 52.863   & 51.930 \\
            & MIT5K\cite{zhao2021deep} & 18.228 & 0.774 & 0.189 & 51.457               & 50.354 \\
            & LOL \cite{zhao2021deep} & 17.947 & 0.822 & 0.272 & 49.334                & 48.334 \\
            & URetinex-Net \cite{wu2022uretinex} & 18.925 & 0.865 & 0.211 & 54.402     & 51.551 \\
            & LOLv1 \cite{xu2022snr} & 15.472 & 0.747 & 0.408 & 26.425                 & 29.915 \\
            & LOLv2real\cite{xu2022snr} & 17.307 & 0.754 & 0.398 & 26.438              & 30.903 \\
            & LOLv2synthetic\cite{xu2022snr} & 17.364 & 0.752 & 0.403 & 23.960         & 29.149 \\ \hline
            
            \multirow{10}{*}{Unsupervised} 
            & Zero-DCE\cite{guo2020zero} & 19.740 & 0.871 & 0.183 & 51.804                        & 51.250 \\
            & Zero-DCE++\cite{li2021learning} & 19.658 & \textcolor{blue}{\underline{0.883}} & 0.182 & 48.573 & 48.435 \\
            & RUAS-LOL\cite{liu2021retinex} & 9.920 & 0.656 & 0.523 & 37.207                      & 40.329 \\
            & RUAS-MIT5K \cite{liu2021retinex}& 13.312 & 0.758 & 0.347 & 45.008                   & 44.523 \\
            & RUAS-DarkFace \cite{liu2021retinex} & 9.696 & 0.642 & 0.517 & 39.655                & 38.934 \\
            & SCI-easy\cite{ma2022toward} & 17.819 & 0.840 & 0.210 & 51.984                       & 50.642 \\
            & SCI-medium\cite{ma2022toward} & 12.766 & 0.762 & 0.347 & 44.176                     & 43.493 \\
            & SCI-difficult\cite{ma2022toward} & 16.993 & 0.837 & 0.232 & 52.369                  & 49.428 \\
            & EnlightenGAN\cite{jiang2021enlightengan}& 17.550 & 0.864 & 0.196 & 48.417           & 48.308 \\
            & ExCNet\cite{zhang2019zero}& 19.437 & 0.865 & 0.168 & 52.576                         & 50.278 \\
            \hline
            
            & Zero-DCE\cite{guo2020zero} & 18.553 & 0.863 & 0.194 & 49.436               & 48.491 \\
            & Zero-DCE++\cite{li2021learning} & 16.018 & 0.832 & 0.240 & 47.253          & 46.000 \\
           Unsupervised  & RUAS\cite{liu2021retinex} & 12.922 & 0.743 & 0.362 & 45.056                & 45.251 \\
            (retrained)& SCI\cite{ma2022toward} & 16.639 & 0.768 & 0.197 & 52.265                   & 51.960 \\
            & EnlightenGAN\cite{jiang2021enlightengan} & 17.957 & 0.849 & 0.182 & 53.871 & 48.261 \\
            & CLIP-LIT\cite{Liang_2023_ICCV}& \textcolor{blue}{\underline{21.579}} & \textcolor{blue}{\underline{0.883}} & \textcolor{blue}{\underline{0.159}} & \textcolor{blue}{\underline{55.682}} & \textcolor{blue}{\underline{52.921}}\\ \hline 
            
            Unsupervised& UBLLIE (Ours) & \textcolor{red}{\textbf{22.017}} & \textcolor{red}{\textbf{0.897}} & \textcolor{red}{\textbf{0.153}}& \textcolor{red}{\textbf{55.691}} &\textcolor{red}{\textbf{53.252}}\\
            \hline 
\end{tabular}
}
\label{tab:baid-table}
\end{table}

\vspace{2mm}\noindent\textbf{Evaluation Metrics}. 
We employ both full-reference and no-reference metrics to evaluate image quality. PSNR and SSIM assess fidelity relative to the ground truth, with PSNR quantifying pixel-level errors and SSIM evaluating luminance, contrast, and structural consistency~\cite{wang2004image}. To better reflect human perception, we also utilize LPIPS, which compares deep feature embeddings and has demonstrated strong alignment with human visual judgments~\cite{zhang2018unreasonable}. Additionally, we report no-reference metrics to assess quality without relying on ground truth. Lightness Order Error (LOE) measures how effectively the lightness order is preserved, which is significant for natural illumination in enhanced images~\cite{wang2013naturalness}. Multi-Scale Structural Similarity (MUS) captures multi-scale structural consistency~\cite{ren2020lr3m}, while the Multi-Scale Image Quality Transformer (MUSIQ) employs a transformer-based model to predict perceptual quality consistent with human assessments~\cite{ke2021musiq}.

\vspace{2mm}\noindent\textbf{Training and Testing Setup}. All experiments are conducted on a workstation equipped with an NVIDIA GeForce RTX 3090 GPU. The framework is implemented in PyTorch. Training comprises three main stages: prompt initialization, initial enhancement training, and prompt refinement via fine-tuning. We use Adam as the optimizer, with a weight decay of 0.001. A cosine annealing learning rate scheduler is employed to stabilize convergence across all phases.

The enhancement network is trained with an initial learning rate of $2 \times 10^{-5}$, while the prompt learning stages utilize a lower learning rate of $5 \times 10^{-6}$. Prompt initialization and refinement are performed over 8k iterations with a batch size of 16, and the enhancement network is trained for 3k epochs with a batch size of 8. A separate validation batch size of four is employed during prompt refinement. Two thresholds are implemented in the prompt learning module: a filtering threshold of 90 to control prompt supervision confidence and an activation threshold of 60 to determine whether prompt guidance is applied. For the image reconstruction module, we use 1k iterations to ensure the enhancement network aligns sufficiently with the refined prompts and structural priors. During inference, the network operates in a feed-forward manner, requiring no paired ground truth, prompt embeddings, or CLIP encoders. This approach ensures efficient deployment while maintaining strong enhancement performance across a wide range of lighting conditions.

\begin{table*}[t]

\centering
\caption{Quantitative comparison of image enhancement performance on LOL~\cite{wei2018deep} and VE-LOL-L~\cite{liu2021benchmarking}. \textcolor{red}{\textbf{Red}} indicates the best result overall, whereas \textcolor{blue}{\underline{blue}} highlights the second-best result among all methods.}
\resizebox{0.9\textwidth}{!}{%
\begin{tabular}{c|c|ccc||ccc}
\hline
 &  & \multicolumn{3}{c||}{LOL~\cite{wei2018deep}} & \multicolumn{3}{c}{VE-LOL-L~\cite{liu2021benchmarking}} \\
Learning&Methods & PSNR $\uparrow$ & SSIM $\uparrow$ & LOE $\downarrow$ & PSNR $\uparrow$ & SSIM $\uparrow$ & LOE $\downarrow$ \\
\hline
\multirow{8}{*}{Supervised }
& LLNet~\cite{lore2017llnet} & 17.91 & 0.76 & 384.21 & 17.38 & 0.73 & 291.59 \\
& LightenNet~\cite{li2018lightennet} & 10.29 & 0.45 & 273.21 & 13.26 & 0.57 & 199.45 \\
& Retinex-Net~\cite{wei2018deep} & 17.24 & 0.55 & 513.28 & 16.41 & 0.64 & 421.41 \\
& MBLLEN~\cite{lv2018mbllen} & 17.90 & 0.77 & 175.10 & 15.95 & 0.70 & 114.91 \\
& KinD~\cite{zhang2019kindling} & 17.57 &\textcolor{blue}{\underline{0.82}} & 377.59 & 18.07 & 0.78 & 253.79 \\
& KinD++~\cite{zhang2021beyond} & 17.60 & 0.80 & 712.12 & 16.80 & 0.74 & 421.97 \\
& TBFEN~\cite{lu2020tbefn} & 17.25 & \textcolor{red}{\textbf{0.83}} & 367.66 & 18.91 & \textcolor{blue}{\underline{0.81}} & 276.65 \\
& DSLR~\cite{lim2020dslr} & 14.98 & 0.67 & 272.68 & 15.70 & 0.68 & 271.63 \\
\hline
\multirow{1}{*}{Semi-supervised }
& DRBN~\cite{yang2020fidelity} & 15.15 & 0.52 & 692.99 & 18.47 & 0.78 & 268.70 \\
\hline
\multirow{6}{*}{Unsupervised }
& EnlightenGAN~\cite{jiang2021enlightengan} & 17.44 & 0.74 & 379.23 & 17.45 & 0.75 & 311.85 \\
& ExCNet~\cite{zhang2019zero} & 16.04 & 0.62 & 220.38 & 16.20 & 0.66 & 225.15 \\
& Zero-DCE~\cite{guo2020zero} & 14.91 & 0.70 & 245.54 & 17.84 & 0.73 & 194.10 \\
& Zero-DCE++~\cite{li2021learning} & 14.86 & 0.62 & 302.06 & 16.12 & 0.45 & 313.50 \\
& RRDNet~\cite{zhu2020zero} & 11.37 & 0.53 & 127.22 & 13.99 & 0.58 & \textcolor{red}{94.23} \\
& CLIP-LIT~\cite{Liang_2023_ICCV} &\color{blue}{\underline{19.33}} &0.73 &\textcolor{blue}{\underline{110.39}}  & \textcolor{blue}{\underline{19.87}}& 0.76& 105.66\\
\hline
Unsupervised  & UBLLIE (Ours) & \textcolor{red}{\textbf{20.97}}&\textcolor{blue}{\underline{0.82}} & \textcolor{red}{\textbf{107.45}}&\textcolor{red}{\textbf{21.02}}&\textcolor{red}{\textbf{0.86}} & \textcolor{blue}{\underline{102.56}} \\
\hline
\end{tabular}%
}
\label{tab:lol-velol-table}
\end{table*}

\section{Comparisons}
\label{sec:results}

\subsection{Quantitative Comparisons}
\label{subsec:quantitative}

We evaluate our model on multiple backlit and low-light datasets, using standard full-reference and no-reference image quality metrics. These metrics include PSNR, SSIM, LPIPS, LOE, and MUSIQ, which collectively measure fidelity, structure, perceptual realism, and naturalness.

As shown in Table~\ref{tab:baid-table}, our method achieves the highest PSNR (22.01), SSIM (0.897), and MUSIQ (55.69) on the BAID test set~\cite{lv2022backlitnet}, while also achieving the lowest LPIPS (0.153), indicating superior visual quality and structural preservation. These results outperform both supervised models, such as KinD++~\cite{zhang2021beyond}, and unsupervised approaches, such as CLIP-LIT~\cite{Liang_2023_ICCV} and Zero-DCE++~\cite{li2021learning}. The last column in Table~\ref{tab:baid-table} presents the results on the Backlit300 dataset~\cite{Liang_2023_ICCV}, for which no ground truth data is available. Our method again achieves the highest MUSIQ score of 55.05, indicating robust perceptual enhancement for real-world, unpaired backlit images.

In the evaluation conducted under low-light conditions, as presented in Table~\ref{tab:lol-velol-table}, our model attains the highest LOE scores on both the LOL dataset~\cite{wei2018deep} (107.45) and the VE-LOL-L dataset~\cite{liu2021benchmarking} (102.56), in addition to yielding competitive PSNR and SSIM outcomes. These findings underscore the model's capability to restore brightness effectively while preserving global lightness consistency and a natural appearance across diverse lighting conditions.

\begin{figure*}[t]
\centering

\begin{subfigure}[t]{0.22\textwidth}
  \includegraphics[width=0.85\linewidth]{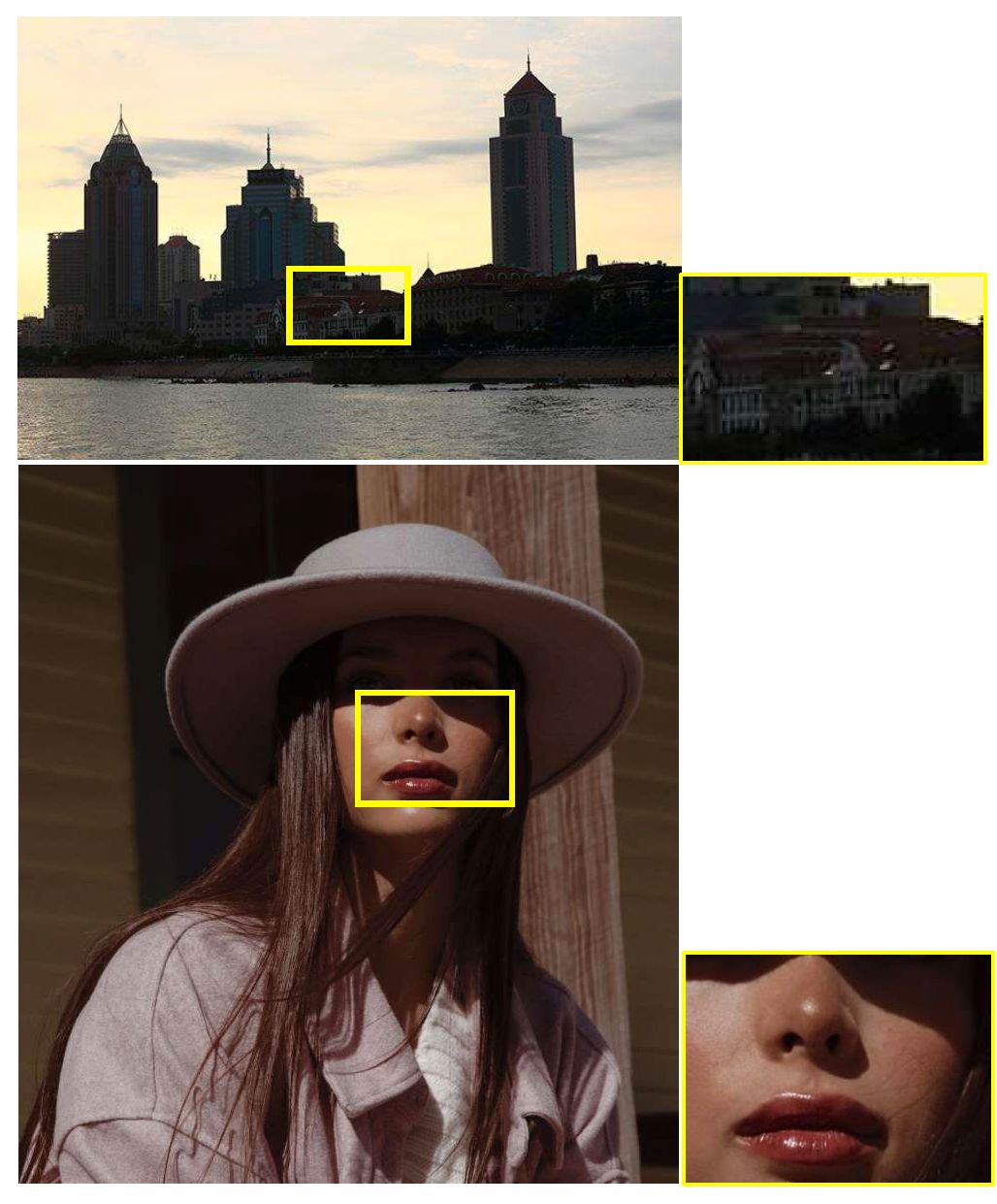}
  \caption*{Input}
\end{subfigure}
\hspace*{1mm}
\begin{subfigure}[t]{0.22\textwidth}
  \includegraphics[width=0.85\linewidth]{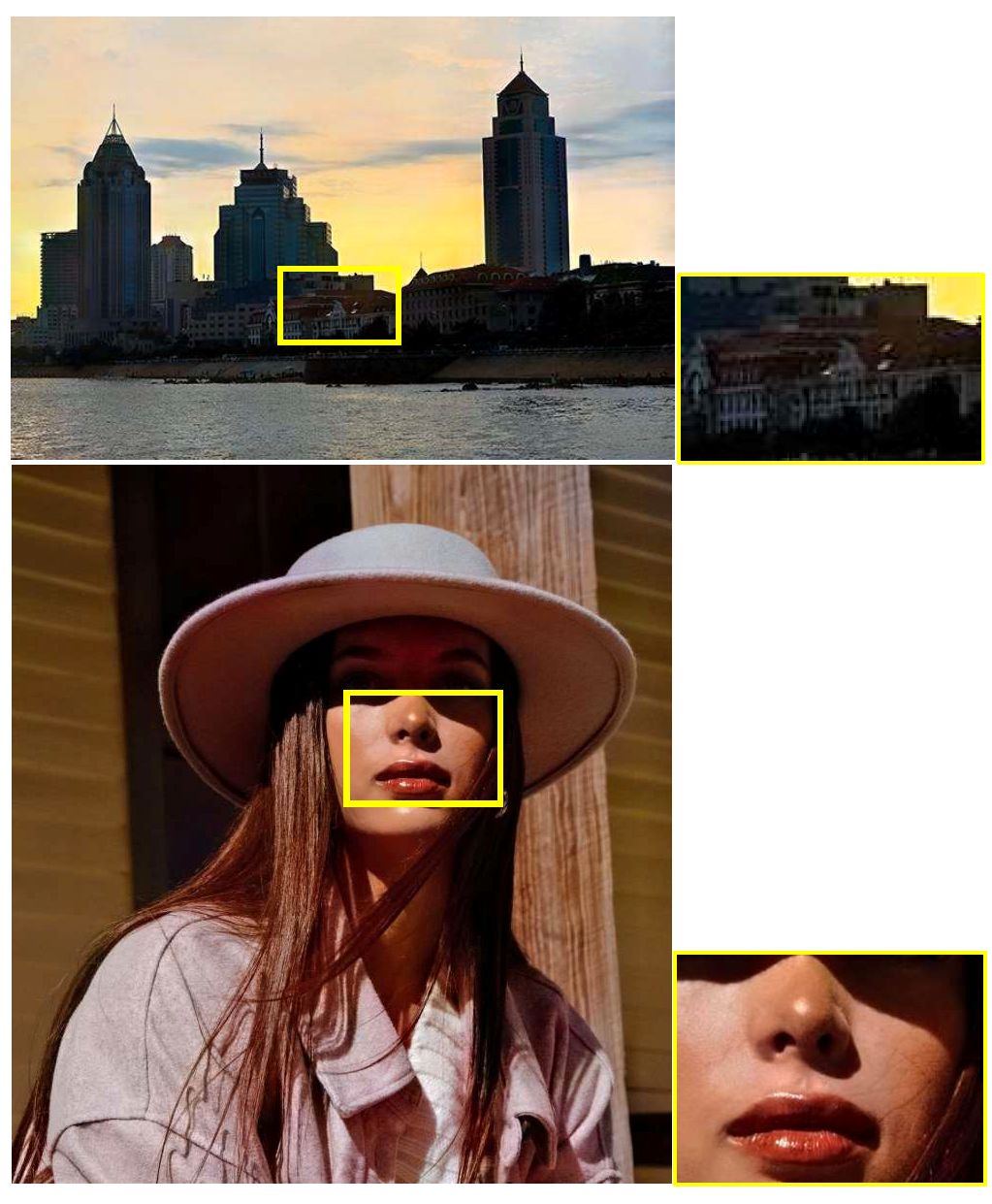}
  \caption*{Afifi~et~al.\ \textnormal{\cite{afifi2021learning}}}
\end{subfigure}
\hspace*{1mm}
\begin{subfigure}[t]{0.22\textwidth}
  \includegraphics[width=0.85\linewidth]{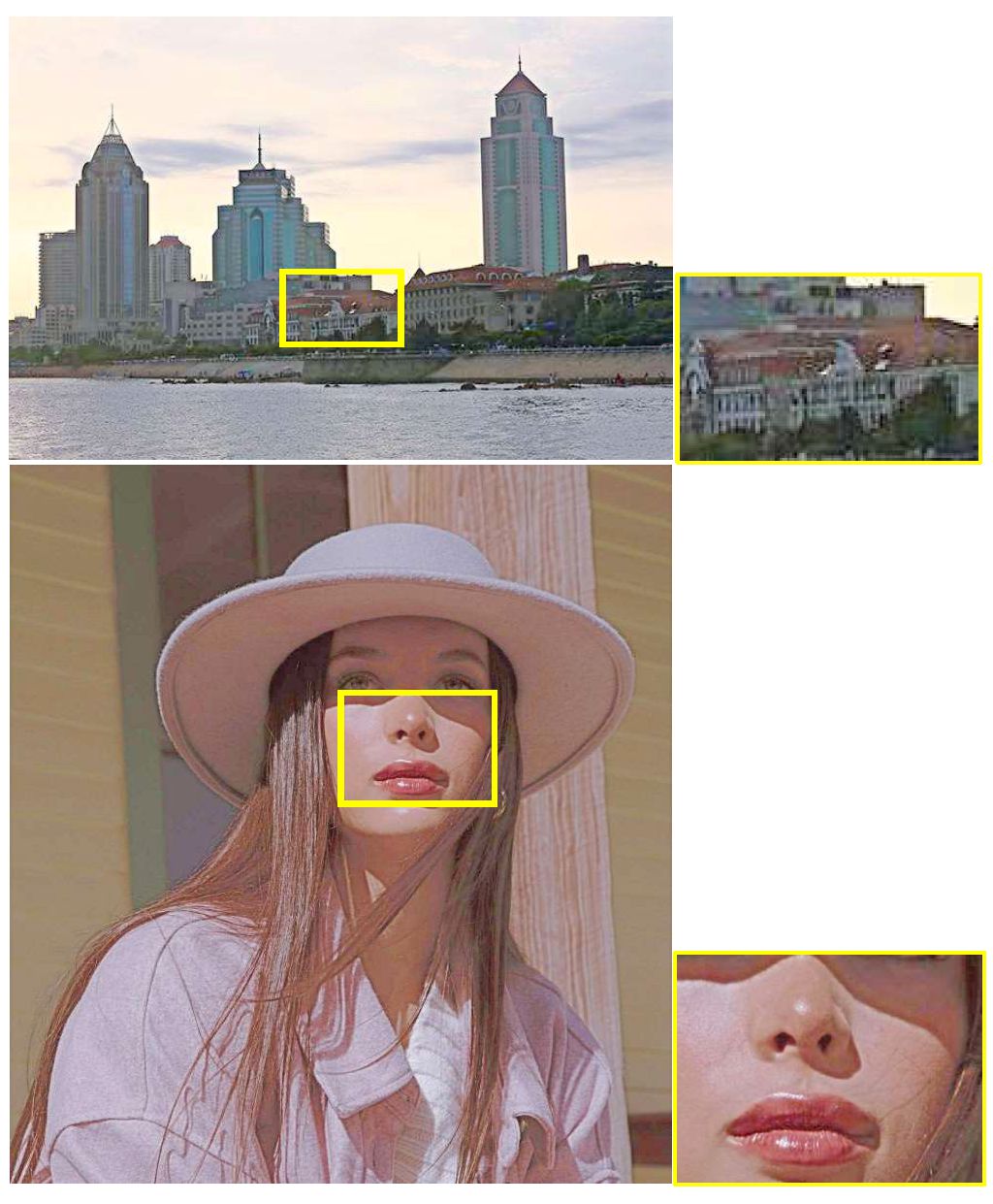}
  \caption*{Zero-DCE \textnormal{\cite{guo2020zero}}}
\end{subfigure}
\hspace*{1mm}
\begin{subfigure}[t]{0.22\textwidth}
  \includegraphics[width=0.85\linewidth]{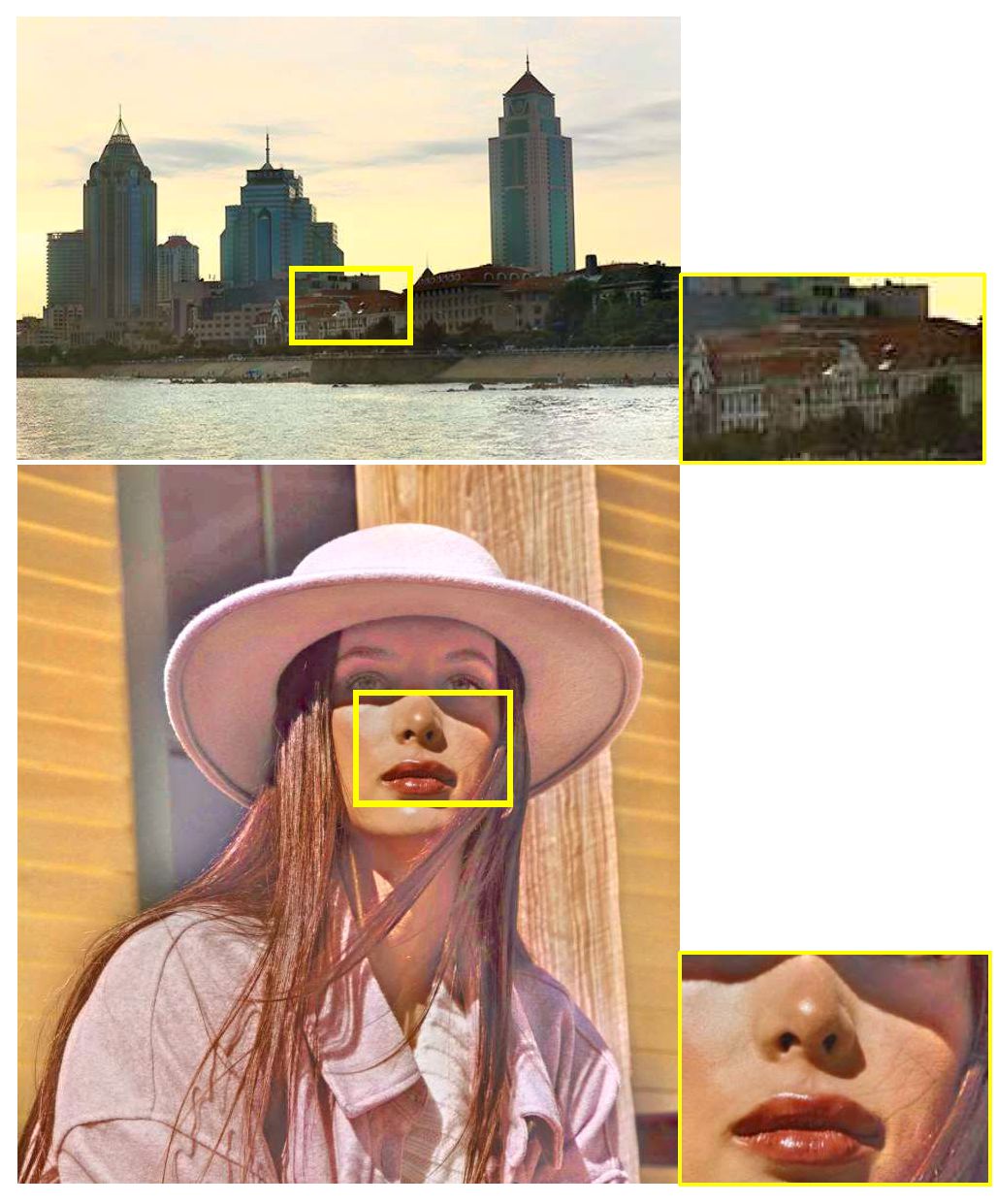}
  \caption*{EnlightenGAN \textnormal{\cite{jiang2021enlightengan}}}
\end{subfigure}

\vspace{3mm}  


\hspace*{3mm}
\begin{subfigure}[t]{0.22\textwidth}
  \includegraphics[width=0.85\linewidth]{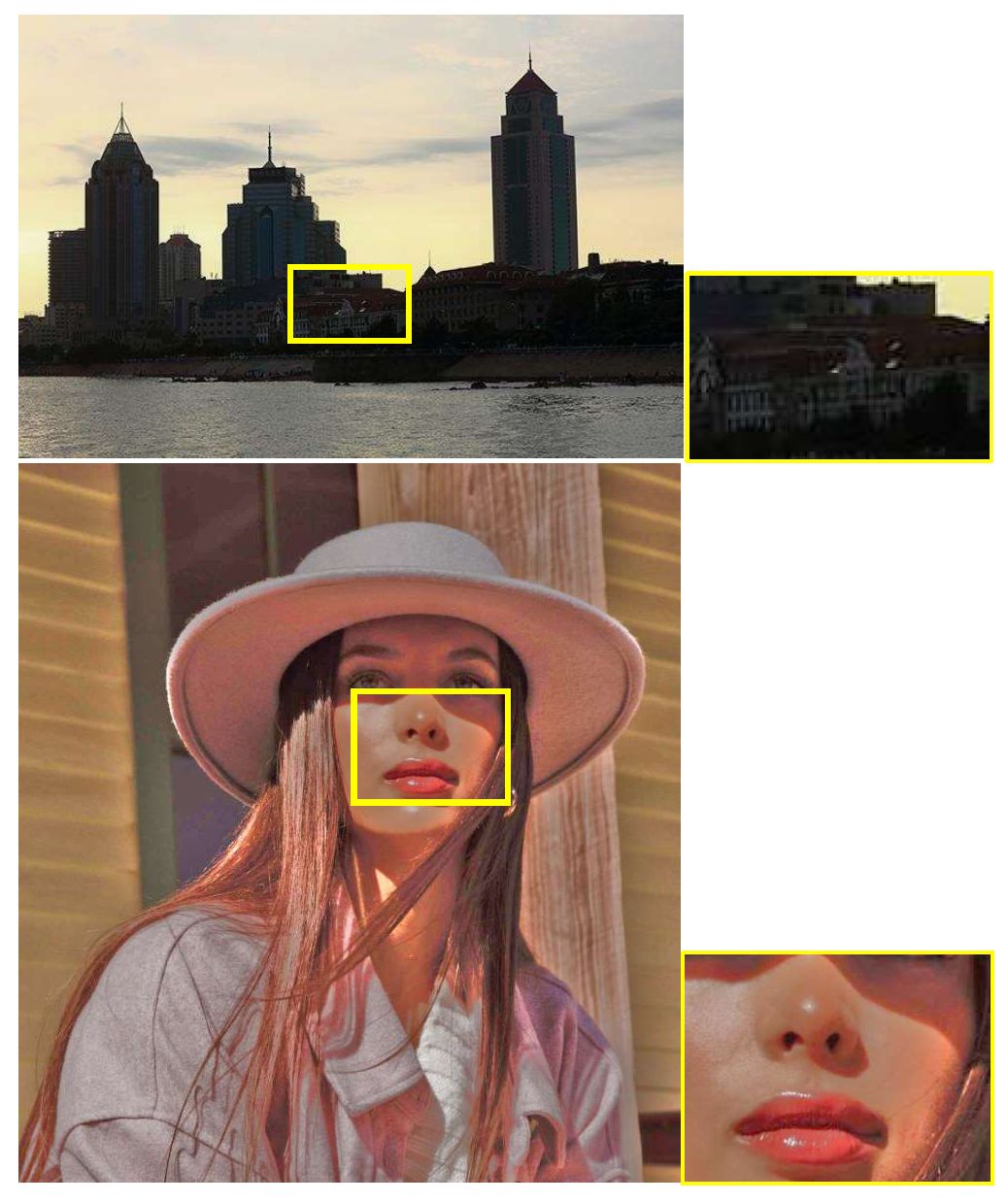}
  \caption*{ExCNet \textnormal{\cite{zhang2019zero}}}
\end{subfigure}
\hspace*{1mm}
\begin{subfigure}[t]{0.22\textwidth}
  \includegraphics[width=0.85\linewidth]{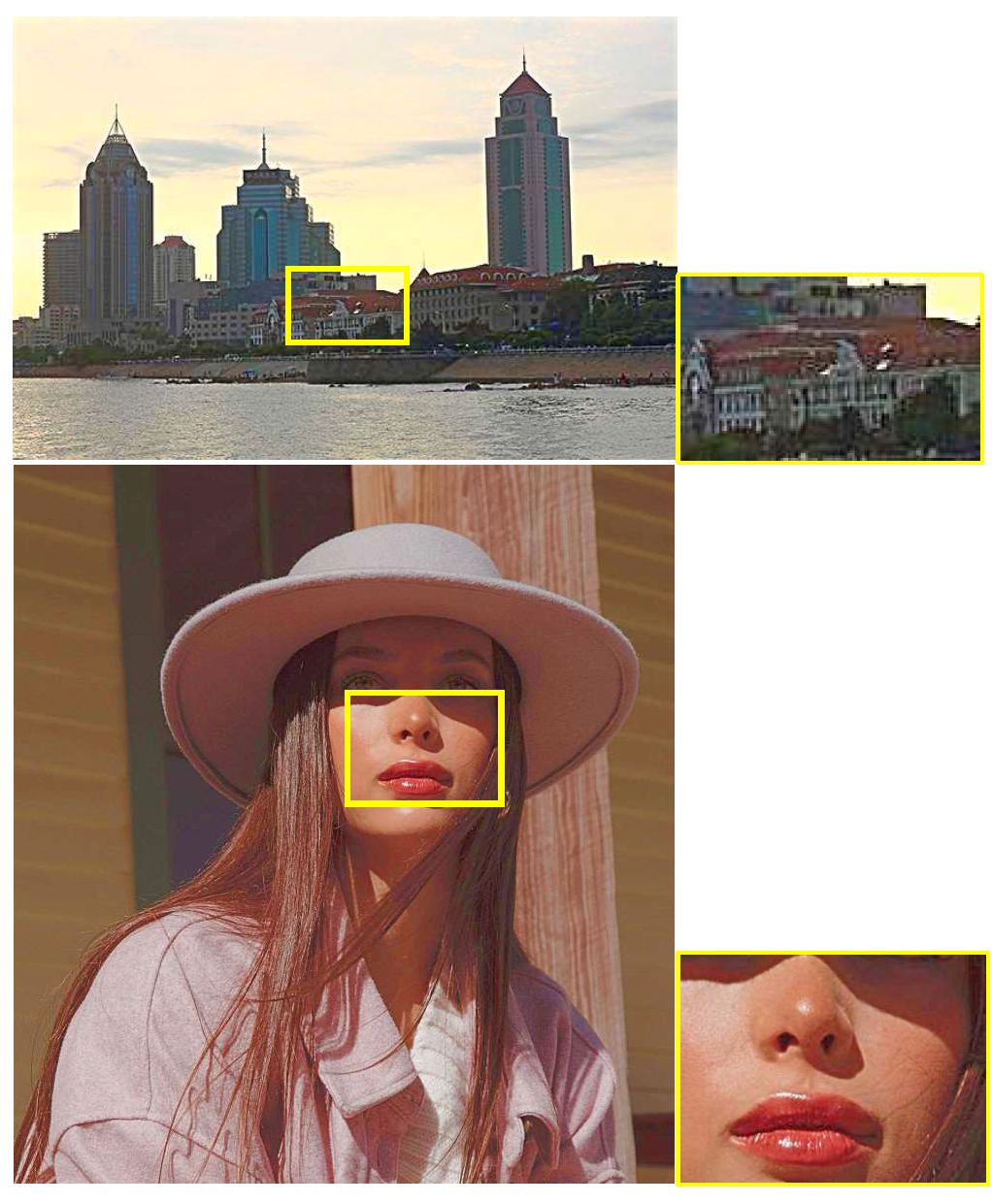}
  \caption*{CLIP-LIT \textnormal{\cite{Liang_2023_ICCV}}}
\end{subfigure}
\hspace*{1mm}
\begin{subfigure}[t]{0.22\textwidth}
  \includegraphics[width=0.85\linewidth]{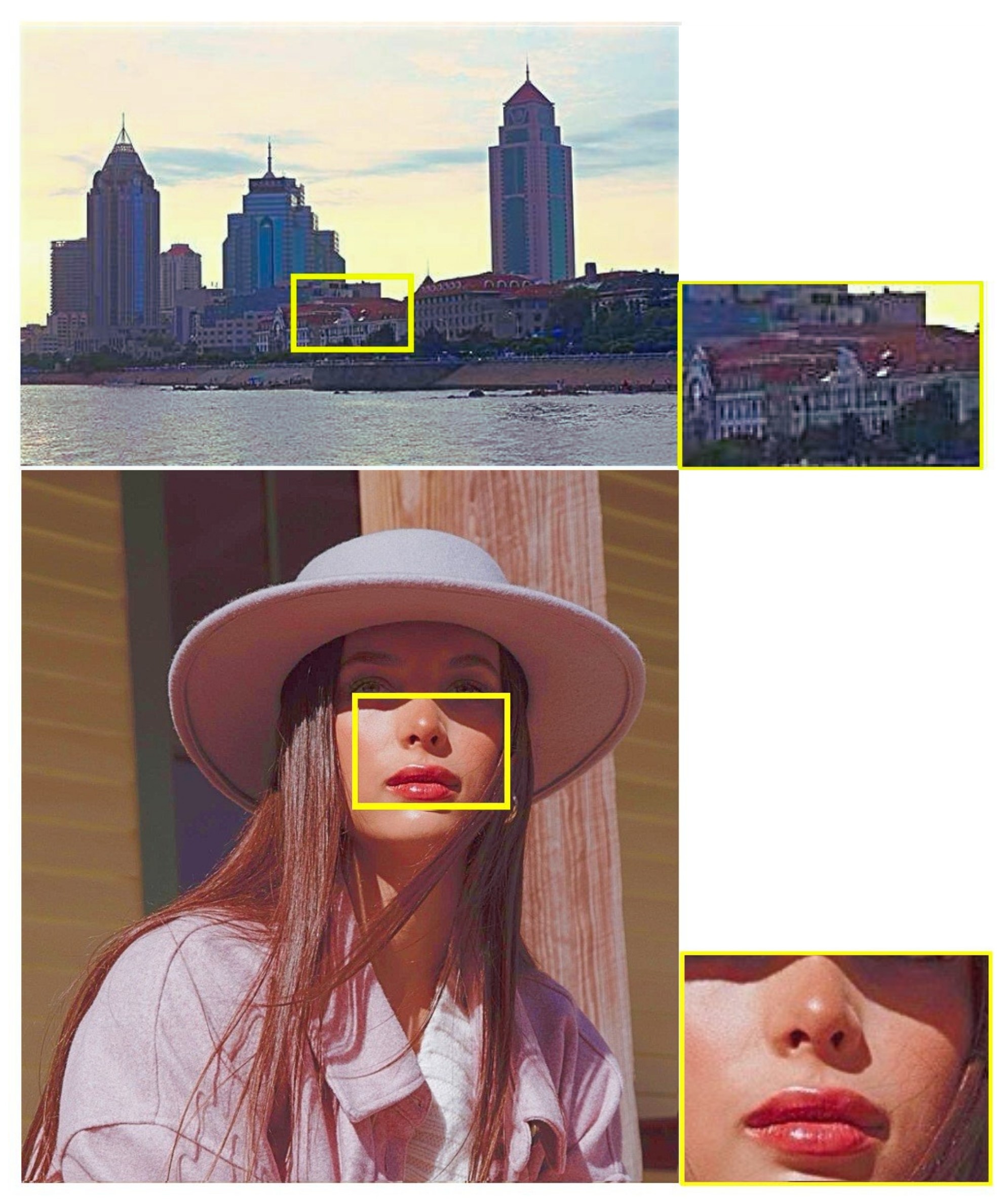}
  \caption*{UBLLIE (Ours)}
\end{subfigure}

\caption{Visual comparison on backlit images sampled from the Backlit300\cite{Liang_2023_ICCV} test dataset. }
\label{fig:backlit300-fig}
\end{figure*}

\subsection{Qualitative Comparisons}
\label{subsec:qualitative}

\begin{figure}[t]
\centering
\includegraphics[width=0.95\linewidth]{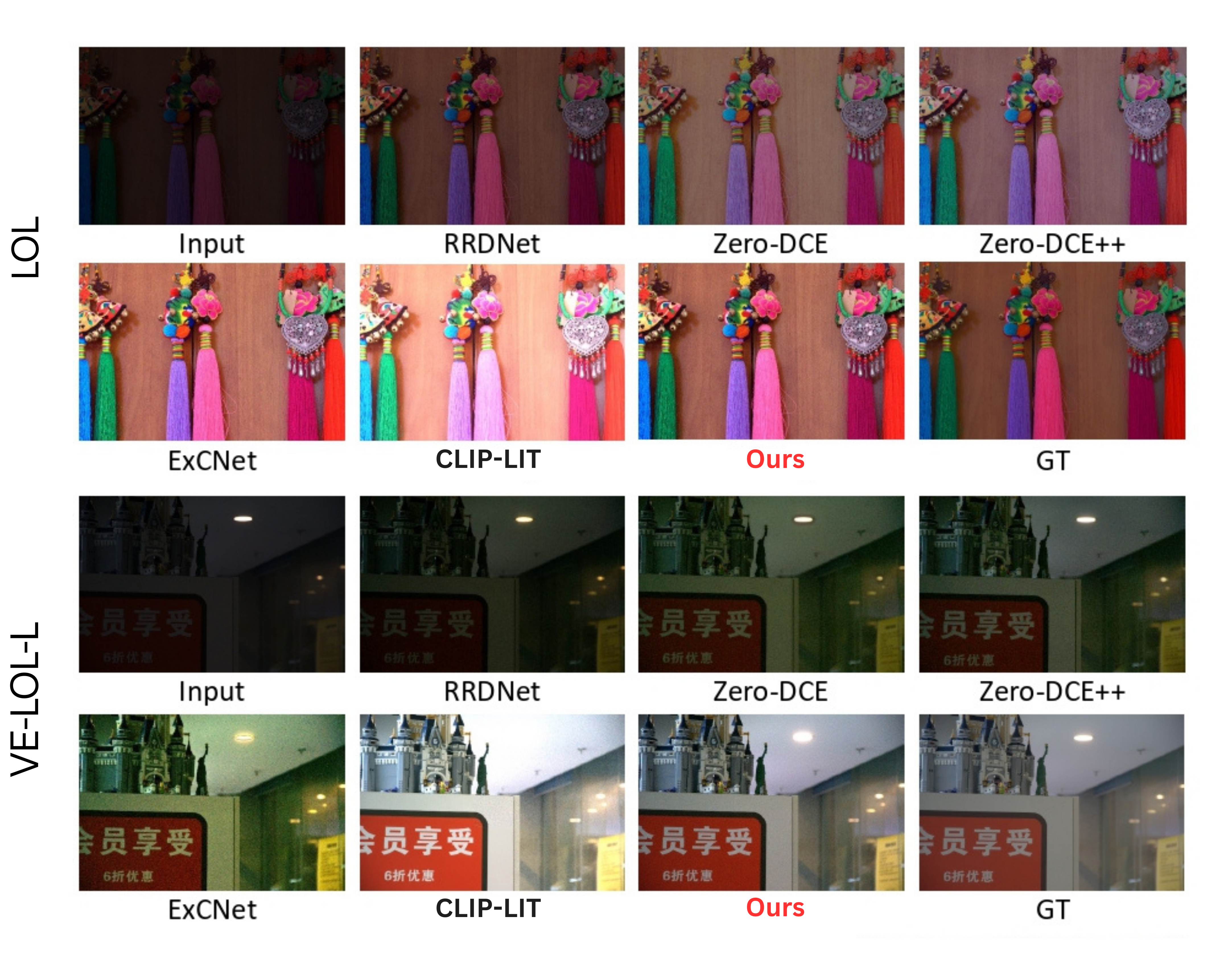}
\caption{Visual comparison of enhancement results on LOL~\cite{wei2018deep} and VE-LOL-L~\cite{liu2021benchmarking} datasets. The top two rows correspond to LOL, and bottom two correspond to VE-LOL-L. Our method (highlighted in red) achieves better visibility and color restoration while maintaining structural integrity, compared to unsupervised and supervised baselines.}
\label{fig:lol_velol_visuals}
\end{figure}

Visual comparisons further confirm the superiority of our method. As illustrated in Figure~\ref{fig:backlit300-fig}, our model generates enhanced results on Backlit300 with improved contrast, detail recovery, and color naturalness compared to existing baselines. Notably, our outputs exhibit none of the typical artifacts, such as haloing or over-brightening, often observed in prior methods. Figure~\ref{fig:backlit_baid} shows examples of enhancement on BAID, where our model produces outputs with better exposure balance across shadow and highlight regions. The images exhibit sharpness and consistency while preserving textures in both the sky and foreground areas. Furthermore, we compare our method to RRDNet~\cite{zhu2020zero}, Zero-DCE++~\cite{li2021learning}, ExCNet, CLIP-LIT~\cite{Liang_2023_ICCV}, and others across the LOL and VE-LOL-L datasets in Figure~\ref{fig:lol_velol_visuals}. Our method (highlighted in red) demonstrates strong brightness recovery, realistic color rendering, and enhanced visibility. Our method retains clearer edges and more faithful color tones under extremely low-light conditions than the unsupervised baselines.

\subsection{Analysis}
\label{subsec:discussion}

The experimental results presented in Tables~\ref{tab:baid-table} and \ref{tab:lol-velol-table} as well as Figures~\ref{fig:backlit300-fig} and \ref{fig:lol_velol_visuals} confirm the effectiveness of our unified framework. Our model achieves consistent improvements across both backlit and low-light datasets under paired and unpaired evaluation protocols.

A key strength lies in CLIP-guided semantic supervision and the enhanced architecture. The symmetric residual U-Net preserves spatial structure, while the ASPP module enables the network to identify illumination inconsistencies across multiple receptive fields. This allows our model to improve underexposed regions without overcompensating for bright areas, addressing a core challenge in both backlit and low-light images. Unlike many existing methods, which are explicitly tailored to either low-light or backlit conditions, our model generalizes well across both, providing a unified, scalable solution. Notably, all results are obtained in an unsupervised setting, without paired ground truth, highlighting the practicality and flexibility of our approach for real-world applications.

\section{Conclusion}
\label{sec:conclusion}

We have introduced a unified framework for the unsupervised enhancement of both backlit and low-light images, merging CLIP-guided semantic supervision with a symmetric residual U-Net enhanced by an ASPP module. Our methodology operates without paired training data and demonstrates strong generalization across diverse lighting conditions. Experiments across multiple benchmarks demonstrate superior performance over both supervised and unsupervised baselines in fidelity, perceptual quality, and structural consistency. Furthermore, this work emphasizes the need for standardized benchmarking for backlit enhancement, an area that remains underexplored compared with low-light scenarios. Our findings on BAID and Backlit300 provide a foundational baseline for future research and underscore the importance of developing robust datasets and comprehensive evaluation protocols for backlit scenes. Although our method is effective, it is constrained by its reliance on a general-purpose CLIP model and its current focus on still images. Future work may explore task-specific vision-language models, temporal extensions for video applications, and lightweight architectures suitable for real-time deployment.

\clearpage  

%
%
\bibliographystyle{splncs04}
\bibliography{refs}
\end{document}